\documentclass{llncs}

\usepackage[utf8]{inputenc}
\usepackage{amsmath}
\usepackage{graphicx}
\usepackage{amsfonts}
\usepackage{amssymb}
\usepackage{url}

\usepackage{todonotes}
\usepackage{booktabs}

\newcommand{\myinstitution}{Macquarie University and the Australian National University}

\newcommand{\institutiona}{Macquarie University}
\newcommand{\institutionb}{the Australian National University}

\begin{document}

{\let\thefootnote\relax\footnotetext{Copyright \textcopyright\ 2020 for this paper by its authors. Use permitted under Creative Commons License Attribution 4.0 International (CC BY 4.0). CLEF 2020, 22-25 September 2020, Thessaloniki, Greece.}}

\title{Query Focused Multi-document Summarisation of Biomedical Texts\thanks{Code associated with this paper is available at \protect\url{https://github.com/dmollaaliod/bioasq8b-public}}}
\subtitle{\myinstitution\ at BioASQ8b}

%\author{Anonymous author 1\inst{1} \and Anonymous author 2\inst{2}}
%\institute{Anonymous Institution 1
%\and
%Anonymous Institution 2}

\author{Diego Moll\'a\orcidID{0000-0003-4973-0963}\inst{1} \and Christopher Jones\orcidID{0000-0002-3491-739X}
\inst{1} \and Vincent Nguyen\orcidID{0000-0003-1787-8090}\inst{2}}
\institute{
Macquarie University, Australia\\
\url{Diego.Molla-Aliod@mq.edu.au}\\
\url{Christopher.Jones4@hdr.mq.edu.au}
\and
Australian National University, Australia\\
\url{Vincent.Nguyen@anu.edu.au} 
}

\maketitle

\begin{abstract}
This paper presents the participation of \myinstitution\ for Task B Phase B of the 2020 BioASQ Challenge (BioASQ8b). Our overall framework implements Query focused multi-document extractive summarisation by applying either a classification or a regression layer to the candidate sentence embeddings and to the comparison between the question and sentence embeddings. We experiment with variants using BERT and BioBERT, Siamese architectures, and reinforcement learning. We observe the best results when BERT is used to obtain the word embeddings, followed by an LSTM layer to obtain sentence embeddings. Variants using Siamese architectures or BioBERT did not improve the results.
\end{abstract}

\section{Introduction}

Query focused multi-document summarisation aims to generate the answer to a question by combining information from multiple documents~\cite{Dang:2006}. This task, therefore, is related to both question answering and text summarisation. There is substantial research in both question answering and text summarisation. In the case of text summarisation, most research focuses on single-document summarisation, and there is also substantial research on multi-document summarisation. However, there is relatively little research on query focused multi-document text summarisation. There are multiple applications where query focused multi-document text summarisation can be useful. A clear example of a useful application is in the domain of biomedicine and clinical medicine, where a doctor or a patient wants to obtain a concise summary of the most relevant evidence related to a particular diagnosis or intervention.

The BioASQ Challenge\footnote{\url{http://www.bioasq.org}} organises shared tasks centered on biomedical texts. The focus of this paper is on the 2020 participation of \myinstitution\ in Task B Phase B (BioASQ8b), where the aim is to find the ``ideal answer'' to a question, given a collection of relevant PubMed abstracts.\footnote{\url{http://www.ncbi.nlm.nih.gov/pubmed}} We approach this task as an instance of query focused multi-document extractive summarisation by scoring each candidate sentence and selecting the top-scoring ones to produce the final summary. \myinstitution\ submitted independent runs to BioASQ8b, but we both used the starting code of Macquarie University's BioASQ7b participation~\cite{Molla:BioASQ7b}. Novel contributions of this paper, compared with previous participation at BioASQ, include:

\begin{enumerate}
    \item The incorporation of BERT and BioBERT in the general architecture of~\cite{Molla:BioASQ7b}.
    \item The use of Siamese architectures in two main setups: 1) sharing weights in the architecture of~\cite{Molla:BioASQ7b}, and 2) using Sentence-BERT \cite{Reimers:2019}.
    \item The use of Proximal Policy Optimisation (PPO) and BERT in a Reinforcement Learning approach.
\end{enumerate}

This paper is structured as follows. Section~\ref{sec:framework} describes the general framework and baselines based on previous participation at BioASQ. Section~\ref{sec:BERT} explains the incorporation of BERT and BioBERT. Section~\ref{sec:Siamese} introduces the incorporation of Siamese architectures. Section~\ref{sec:RL} describes the use of Reinforcement Learning. Section~\ref{sec:results} presents and discusses the results of cross-validation evaluations using the BioASQ8b training data. Section~\ref{sec:submissions} presents and discusses the runs submitted to BioASQ. Finally, Section~\ref{sec:conclusions} concludes this paper.

%\section{Related Work}

\section{General Framework and Baselines}\label{sec:framework}

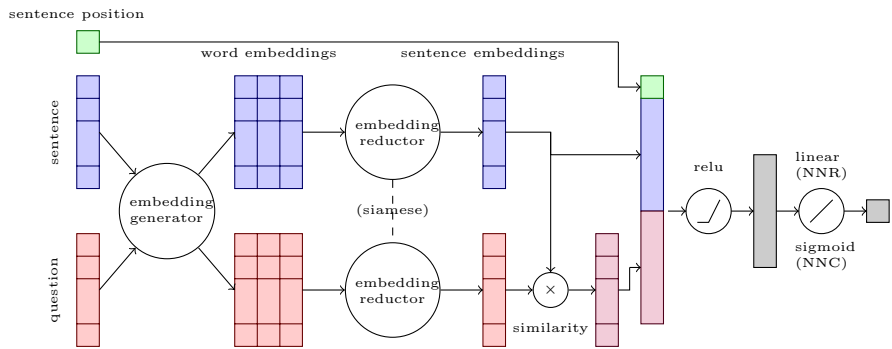
\begin{figure}
  \centering
    \begin{tikzpicture}[scale=0.3]
    \tiny
% input
    \filldraw[fill=blue!20!white, draw=blue!40!black] (0,0) rectangle (1,5) (0,1) -- (1,1) (0,3) -- (1,3) (0,4) -- (1,4);
    \filldraw[fill=red!20!white, draw=red!40!black] (0,-7) rectangle (1,-2) (0,-6) -- (1,-6) (0,-4) -- (1,-4) (0,-3) -- (1,-3);
    \draw (-1,2.5) node[rotate=90] {sentence};
    \draw (-1,-4.5) node[rotate=90] {question};

% word embeddings
    \draw (4,-1) node [circle,draw,align=center,text width=1cm] (em) {embedding generator};
    \draw (8.5,6) node {word embeddings};
    \filldraw[fill=blue!20!white, draw=blue!40!black] (7,0) rectangle (10,5) (7,1) -- (10,1) (7,3) -- (10,3) (7,4) -- (10,4) (8,0) -- (8,5) (9,0) -- (9,5);
    \filldraw[fill=red!20!white, draw=red!40!black] (7,-7) rectangle (10,-2) (7,-6) -- (10,-6) (7,-4) -- (10,-4) (7,-3) -- (10,-3) (8,-7) -- (8,-2) (9,-7) -- (9,-2);

    \draw[->] (1,2.5) -- (em);
    \draw[->] (1,-4.5) -- (em);

    \draw[->] (em) -- (7,2.5);
    \draw[->] (em) -- (7,-4.5);
% sentence embeddings
    \draw (14,2.5) node [circle,draw,align=center,text width=1cm] (sr) {embedding reductor};
    \draw (14,-4.5) node [circle,draw,align=center,text width=1cm] (qr) {embedding reductor};
    \draw[dashed] (sr) -- node {(siamese)} (qr);
    \draw (18,6) node {sentence embeddings};
    \filldraw[fill=blue!20!white, draw=blue!40!black] (18,0) rectangle (19,5) (18,1) -- (19,1) (18,3) -- (19,3) (18,4) -- (19,4);
    \filldraw[fill=red!20!white, draw=red!40!black] (18,-7) rectangle (19,-2) (18,-6) -- (19,-6) (18,-4) -- (19,-4) (18,-3) -- (19,-3);

    \draw[->] (10,2.5) -- (sr);
    \draw[->] (sr) -- (18,2.5);
    \draw[->] (10,-4.5) -- (qr);
    \draw[->] (qr) -- (18,-4.5);

% similarity
    \draw (21,-4.5) node [circle,draw] (t) {$\times$};
    \filldraw[fill=purple!20!white, draw=purple!40!black] (23,-7) rectangle (24,-2) (23,-6) -- (24,-6) (23,-4) -- (24,-4) (23,-3) -- (24,-3);

    \draw[->] (19,2.5) -| (t);
    \draw[->] (19,-4.5) -- (t);

    \draw[->] (21,2.5) |- (25,1.5);
    \draw[->] (t) -- (23,-4.5);
    \draw (24,-4.5) -| (24.5,-3.5);
    \draw[->] (24.5,-3.5) -- (25,-3.5);

    \draw (21,-6.2) node {similarity};

% hidden layer
    \filldraw[fill=blue!20!white, draw=blue!40!black] (25,-1) rectangle (26,4);
    \filldraw[fill=purple!20!white, draw=purple!40!black] (25,-1) rectangle (26,-6);
    \draw (28,-1) circle[radius=1] (27.5,-1.5) -- (28,-1.5) -- (28.5,-0.5);
    \draw (28,1) node {relu};

    \filldraw[fill=black!20!white, draw=black] (30,-3.5) rectangle (31,1.5);

    \draw[->] (26.2,-1) -- (27,-1);
    \draw[->] (29,-1) -- (30,-1);

% final layer
    \draw (33,-1) circle[radius=1] (32.5,-1.5) -- (33.5,-0.5);
    \filldraw[fill=black!20!white, draw=black] (35,-1.5) rectangle (36,-0.5);
    \draw (33.5,1) node[text width=1cm] {linear (NNR)};
    \draw (33.5,-3) node[text width=1cm] {sigmoid (NNC)};

    \draw[->] (31,-1) -- (32,-1);
    \draw[->] (34,-1) -- (35,-1);
    
% position
    \filldraw[fill=green!20!white, draw=green!40!black] (0,6) rectangle (1,7);
    \draw (1,6.5) -- (24,6.5);
    \draw[->] (24,6.5) |- (25,4.5);
    \filldraw[fill=green!20!white, draw=green!40!black] (25,5) rectangle (26,4);
    \draw (0,7.7) node {sentence position};
  \end{tikzpicture}
  \caption{Base architecture. An embedding generator calculates the embeddings of each of the words of the candidate sentence and the question. These are then passed to an embedding reductor that obtains the overall sentence embeddings of the candidate sentence and the question. A Siamese variant enforced shared weights of the embedding reductors. Then, the sentence position is concatenated with the sentence embedding and the similarity of sentence and question embeddings, implemented as a product. A final layer predicts the label of the sentence.}
  \label{fig:architecture}
\end{figure}

The overall architecture of our systems is based on that of Macquarie University's participation to BioASQ7b~\cite{Molla:BioASQ7b}. This architecture is described in Fig.~\ref{fig:architecture}.
The baseline systems re-use the following options from~\cite{Molla:BioASQ7b} to generate the embeddings of the words and sentences.

\begin{itemize}
    \item The \textbf{embedding generator} to obtain word embeddings is a matrix of pre-trained embeddings generated by word2vec. We trained word2vec using PubMed documents provided by the organisers of BioASQ. The embeddings had a vector size of~100.
    \item The \textbf{embedding reductor} to obtain sentence embeddings is a pair of forward and backward LSTM chains. The weights of the embedding reductor for the candidate sentence and the question were not shared.
    \item The activation function of the \textbf{final layer} is a linear function for the regression setup  and a sigmoid function for the classification setup.
    \item The similarity between the embeddings of the candidate sentence and the question is the element-wise product.
\end{itemize}

As in \cite{Molla:BioASQ7b}, we tried a classification setup and a regression setup. In the regression setup (``NNR'' in Fig.~\ref{fig:architecture} Table~\ref{tab:experiments}), the training data is labelled with the ROUGE-SU4 F1 value of the candidate sentence and the objective function to optimise is the Mean Squared Error. In the classification setup (``NNC'' in Fig.~\ref{fig:architecture} Table~\ref{tab:experiments}), the~5 candidate sentences with the highest ROUGE-SU4 F1 are labelled as~1, and the rest are labelled as~0. The objective function to optimise in the classification setup is binary cross-entropy.

In both the classification and the regression setup, the summary is produced by scoring each candidate sentence and extracting the top $n$ sentences to generate the summary, where $n$ depended on the question type and was the same as reported in~\cite{Molla:BioASQ7b} (Table~\ref{tab:n}).

\begin{table}[]
    \centering
    \caption{Number of sentences selected, for each question type}
    \label{tab:n}
    \begin{tabular}{ccccc}
    & \textbf{Summary} & \textbf{Factoid} & \textbf{Yesno} & \textbf{List} \\
    \midrule
    \textbf{n}     &  6 & 2 & 2 & 3\\
    \end{tabular}
\end{table}

\section{Experiments with BERT and BioBERT}\label{sec:BERT}

BERT~\cite{Devlin:2018} has been used in a wide range of NLP tasks, including text classification~\cite{mekala:2020}, and extractive~\cite{liu:2019} and abstractive~\cite{lewis:2020} summarisation. We have integrated BERT into our general architecture of Fig.~\ref{fig:architecture} as described below.

In a first experiment (``BERT untrained'' in Table~\ref{tab:experiments}), we replaced the embedding generator with BERT using the pre-trained model provided by Huggingface.\footnote{\url{https://huggingface.co/}\label{foot:huggingface}} The resulting word embeddings are now affected by context. The BERT weights were not updated during the training stage. Following the recommendation of the Huggingface library, the embedding reductor of each candidate sentence and the question are the average of the word embeddings.

In a subsequent experiment (``BERT trained'' in Table~\ref{tab:experiments}), the embedding generator and reductors are as in BERT untrained, but we allowed the BERT weights to be fine-tuned during the training process.

We also tried a variant (``BERT LSTM'' in Table~\ref{tab:experiments}) that uses BERT in the embedding generator as in BERT untrained, but the embedding reductor is a bidirectional LSTM chain as in the NNC baseline. This variant is therefore comparable with the NNC baseline, and the only difference being the use of BERT for the embedding generator.

In a final series of experiments (``BioBERT untrained'' and ``BioBERT LSTM'' in Table~\ref{tab:experiments}), the embedding generator and reductors are as in BERT untrained and BERT LSTM, but the pre-trained model was as provided by the developers of BioBERT\footnote{\url{https://github.com/dmis-lab/biobert}} \cite{Lee:2019}, who used biomedical documents to pre-train BERT.

In all of the experiments in this section, the final layer was a classification layer with a sigmoid activation and the objective function to optimise was binary cross-entropy.

\section{Experiments with Siamese Networks}\label{sec:Siamese}

Siamese networks have been used in applications that include a comparison between documents, for example to determine semantic similarity \cite{Reimers:2019}. The general idea is to use the same processing module for each of the two documents by sharing the weights of the encoding component that generates the embeddings of the documents. We have used this idea in two main kinds of experiments described below.

\subsection{Siamese LSTM}\label{sec:SiamLSTM}

A straightforward implementation of Siamese Networks (``Siamese LSTM'' in Table~\ref{tab:experiments}) using the overall architecture of Fig.~\ref{fig:architecture} shares the weights of the embedding reductors of the candidate sentence and the question. This ensures that the sentence embeddings are generated using exactly the same process.

\subsection{Sentence-BERT}\label{sec:SiamBERT}

\begin{figure}
  \centering
    \begin{tikzpicture}[scale=0.3]
    \tiny
% input
    \filldraw[fill=blue!20!white, draw=blue!40!black] (0,0) rectangle (1,5) (0,1) -- (1,1) (0,3) -- (1,3) (0,4) -- (1,4);
    \filldraw[fill=red!20!white, draw=red!40!black] (0,-7) rectangle (1,-2) (0,-6) -- (1,-6) (0,-4) -- (1,-4) (0,-3) -- (1,-3);
    \draw (-1,2.5) node[rotate=90] {sentence};
    \draw (-1,-4.5) node[rotate=90] {question};

% sentence embeddings
    \draw (4,-1) node [circle,draw,align=center,text width=1cm] (em) {SBERT};
    \draw (8.5,6) node {sentence embeddings};
    \filldraw[fill=blue!20!white, draw=blue!40!black] (8,0) rectangle (9,5) (8,1) -- (9,1) (8,3) -- (9,3) (8,4) -- (9,4);
    \filldraw[fill=red!20!white, draw=red!40!black] (8,-7) rectangle (9,-2) (8,-6) -- (9,-6) (8,-4) -- (9,-4) (8,-3) -- (9,-3);

    \draw[->] (1,2.5) -- (em);
    \draw[->] (em) -- (8,2.5);
    \draw[->] (1,-4.5) -- (em);
    \draw[->] (em) -- (8,-4.5);

% regression layer
    \draw (11,-1) node [circle,draw] (c) {$\cos$};
    \filldraw[fill=black!20!white, draw=black] (14,-1.5) rectangle (15,-0.5);
    \draw (14.5,0) node {regr};

    \draw[->] (9,2.5) -- (c);
    \draw[->] (9,-4.5) -- (c);
    \draw[->] (c) -- (14,-1);

% difference layer
    \draw (17,-1) node [circle,draw] (d) {$|-|$};
    \draw (20.5,2.5) node {difference};
    \draw[->] (9,2.5) -| (d);
    \draw[->] (9,-4.5) -| (d);

    \filldraw[fill=purple!20!white, draw=purple!40!black] (20,-3.5) rectangle (21,1.5) (20,-2.5) -- (21,-2.5) (20,-0.5) -- (21,-0.5) (20,0.5) -- (21,0.5);
    
    \draw[->] (17,2.5) |- (25,4);
    \draw[->] (d) -- (20,-1);
    \draw[->] (17,-4.5) |- (25,-6);
    \draw[->] (21,-1) -- (25,-1);

% classification layer
    \filldraw[fill=blue!20!white, draw=blue!40!black] (25,1.5) rectangle (26,6.5);
    \filldraw[fill=purple!20!white, draw=purple!40!black] (25,-3.5) rectangle (26,1.5);
    \filldraw[fill=red!20!white, draw=red!40!black] (25,-8.5) rectangle (26,-3.5);
    \draw (29,-1) node[circle,draw] (s) {$\int$};% (28.5,-1.5) -- (29.3,-0.5);
    \draw (29,0.5) node {softmax};

    \filldraw[fill=black!20!white, draw=black] (32,-1.5) rectangle (33,-0.5);
    \draw (32.5,0) node {clas};
    \draw[->] (26,-1) -- (s);
    \draw[->] (s) -- (32,-1);
    
% position
%    \filldraw[fill=green!20!white, draw=green!40!black] (0,6) rectangle (1,7);
%    \draw (1,6.5) -- (24,6.5);
%    \draw[->] (24,6.5) |- (25,4.5);
%    \filldraw[fill=green!20!white, draw=green!40!black] (25,5) rectangle (26,4);
%    \draw (0,7.7) node {sentence position};
  \end{tikzpicture}
  \caption{Architecture for the experiments with SBERT. SBERT is used to generate the embeddings of the candidate sentence and the question. The embeddings are then processed in either a regression setup or a classification setup.}
  \label{fig:sbert}
\end{figure}
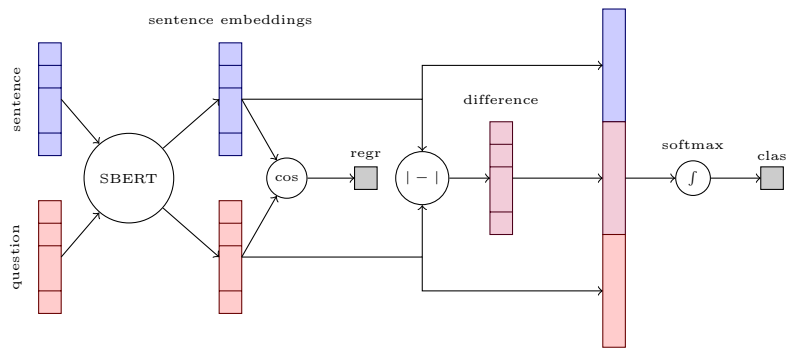

The second implementation of Siamese Networks uses Sentence-BERT (SBERT)~\cite{Reimers:2019} to determine whether a candidate sentence is similar to the question. In particular, the system uses BERT \cite{Devlin:2018} in a Siamese setup as described in Fig.~\ref{fig:sbert}.

Depending on the settings, the system uses regression, classification or both (multi-task), for prediction. 

\begin{itemize}
    \item In the \textbf{regression} setup, the cosine similarity between candidate sentence and question embeddings is computed, and the objective function to optimise is the Mean Squared Error (MSE).

    \item In the \textbf{classification} setup, the candidate sentence and question embeddings are concatenated with the element-wise absolute difference between the sentence and question, and a softmax layer is added. The objective function to optimise is binary cross-entropy.

    \item In the \textbf{multi-task} setup, the classification and regression setups are jointly optimised during training. At prediction time, we use either the classification head or the regression head.
\end{itemize}

In both the regression and the classification setups, the training data was labelled with the classification labels mentioned in Section~\ref{sec:framework}.

We also experimented with a BERT and a BioBERT variation\footnote{\url{https://huggingface.co/gsarti/biobert-nli}} of a Sentence Transformer~\cite{Reimers:2019} to ensure that the sentence embeddings reflect a biomedical word-space. 

Table~\ref{tab:sbert_names} shows all the combinations we tried.
\begin{table}[]
    \centering
    \caption{Variants of SBERT used in our experiments}\label{tab:sbert_names}
    \begin{tabular}{llcccc}
    \toprule
\textbf{System Name}        & \textbf{Model} & \multicolumn{2}{c}{\textbf{Training}} & 
\multicolumn{2}{c}{\textbf{Prediction}}\\
&&\textbf{Classification} & \textbf{Regression} & \textbf{Classification} & \textbf{Regression}\\
\midrule
SBERT R         & BERT & & Y & & Y\\
SBERT C         & BERT & Y & & Y &\\
SBERT M R         & BERT & Y & Y & & Y\\
SBERT M C         & BERT & Y & Y & Y &\\
\midrule
SBioBERT R         & BioBERT & & Y & & Y\\
SBioBERT C         & BioBERT & Y & & Y &\\
SBioBERT M R         & BioBERT & Y & Y & & Y\\
SBioBERT M C         & BioBERT & Y & Y & Y &\\
\bottomrule
    \end{tabular}
    \label{tab:my_label}
\end{table}

% [1] BERT
% [2] https://www.aclweb.org/anthology/D19-1410/

\section{Experiments with Reinforcement Learning}\label{sec:RL}

Reinforcement learning allows the training process to optimise the target evaluation metric (ROUGE-SU4 F1) directly. Whereas~\cite{Molla:BioASQ7b} used the REINFORCE algorithm~\cite{williams:1992}, in our participation to BioASQ8b we used the Proximal Policy Optimisation (PPO) approach.
We choose PPO for our summarisation task because past research shows that it penalises changes to the policy \cite{schulman:2017}, and because we observe a more consistent learning curve using this approach compared to REINFORCE~\cite{jones:2020}.

\subsection{Approach}\label{sec:ReApproach}
As in past submissions, our reinforcement learning system classifies sentences to be either (0) not included in the summary or (1) included in the summary based on a policy, using the ROUGE-SU4 F1 score directly as the reward.
We use the stable baselines library\footnote{\url{https://stable-baselines.readthedocs.io/}\label{foot:stable}} \cite{stablebaselines} to implement the PPO reinforcement learning approach, and apply this to our BioASQ summarisation task environment.

We perform some hyperparameter tuning, while leaving the PPO code unmodified as much as possible.
We use a horizon ($n$ steps) of 1000 with 4 mini-batches each, and run for a total of 500,000 timesteps.
When training our model, we choose any action from the probability distribution of the policy function, but when testing we choose the action out of 100 samples.

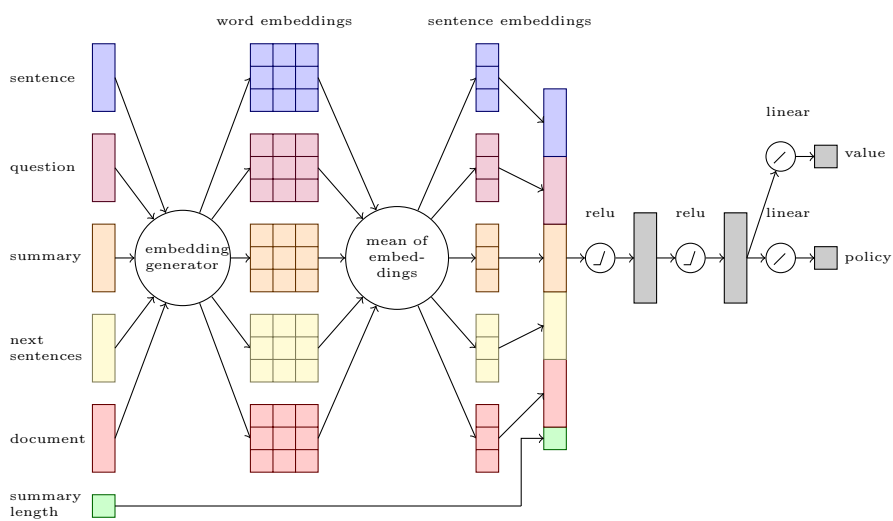
\begin{figure}
  \begin{center}
\tikzstyle{decision} = [diamond, draw, fill=blue!20, 
    text width=4.5em, text badly centered, node distance=3cm, inner sep=0pt]
\tikzstyle{block} = [rectangle, draw, fill=blue!20, 
    text width=5em, text centered, rounded corners, minimum height=4em]
\tikzstyle{line} = [draw, -latex']
\tikzstyle{cloud} = [draw, ellipse,fill=red!20, node distance=3cm,
    minimum height=2em]
    \begin{tikzpicture}[scale=0.3]
    \tiny
    \filldraw[fill=red!20!white, draw=red!40!black] (0,0) rectangle (1,3) ; %INPUT
    \filldraw[fill=yellow!20!white, draw=yellow!40!black] (0,4) rectangle (1,7) ; 
    \filldraw[fill=orange!20!white, draw=orange!40!black] (0,8) rectangle (1,11) ; 
    \filldraw[fill=purple!20!white, draw=purple!40!black] (0,12) rectangle (1,15); 
    \filldraw[fill=blue!20!white, draw=blue!40!black] (0,16) rectangle (1,19); 
    \filldraw[fill=green!20!white, draw=green!40!black] (0,-1) rectangle (1,-2) ;
    \draw (-2,17.5) node[rotate=0,text width=1cm] {sentence};
    \draw (-2,13.5) node[rotate=0,text width=1cm] {question};
    \draw (-2,9.5) node[rotate=0, text width=1cm] {summary};
    \draw (-2,5.5) node[rotate=0, text width=1cm] {next \hbox{sentences}};
    \draw (-2,1.5) node[rotate=0, text width=1cm] {document};
    \draw (-2,-1.5) node[rotate=0,text width=1cm] {summary length};
    \draw (4,9.5) node [circle,draw,align=center,text width=1cm] (em) {embedding generator};
    \draw[->] (1,1.5) -- (em);
    \draw[->] (1,5.5) -- (em);
    \draw[->] (1,9.5) -- (em);
    \draw[->] (1,13.5) -- (em);
    \draw[->] (1,17.5) -- (em);
    \draw[->] (em) -- (7,1.5);
    \draw[->] (em) -- (7,5.5);
    \draw[->] (em) -- (7,9.5);
    \draw[->] (em) -- (7,13.5);
    \draw[->] (em) -- (7,17.5);
    \draw (8.5,20) node[rotate=0] {word embeddings}; %word level
    \filldraw[fill=red!20!white, draw=red!40!black] (7,0) rectangle (10,3) (7,1) -- (10,1) (7,2) -- (10,2) (8,0) -- (8,3) (9,0) -- (9,3);
    \filldraw[fill=yellow!20!white, draw=yellow!40!black] (7,4) rectangle (10,7) (7,5) -- (10,5) (7,6) -- (10,6) (8,4) -- (8,7) (9,4) -- (9,7);
    \filldraw[fill=orange!20!white, draw=orange!40!black] (7,8) rectangle (10,11) (7,9) -- (10,9) (7,10) -- (10,10) (8,8) -- (8,11) (9,8) -- (9,11);
    \filldraw[fill=purple!20!white, draw=purple!40!black] (7,12) rectangle (10,15) (7,13) -- (10,13) (7,14) -- (10,14) (8,12) -- (8,15) (9,12) -- (9,15);
    \filldraw[fill=blue!20!white, draw=blue!40!black] (7,16) rectangle (10,19) (7,17) -- (10,17) (7,18) -- (10,18) (8,16) -- (8,19) (9,16) -- (9,19);
    \draw (13.5,9.5) node [circle,draw,align=center,text width=1cm] (mean) {mean of embeddings};
    \draw[->] (10,1.5) -- (mean);
    \draw[->] (10,5.5) -- (mean);
    \draw[->] (10,9.5) -- (mean);
    \draw[->] (10,13.5) -- (mean);
    \draw[->] (10,17.5) -- (mean);
    \draw[->] (mean) -- (17,1.5);
    \draw[->] (mean) -- (17,5.5);
    \draw[->] (mean) -- (17,9.5);
    \draw[->] (mean) -- (17,13.5);
    \draw[->] (mean) -- (17,17.5);
    \draw (18.5,20) node[rotate=0] {sentence embeddings}; %sentence level
    \filldraw[fill=red!20!white, draw=red!40!black] (17,0) rectangle (18,3) (17,1) -- (18,1) (17,2) -- (18,2) ;
    \filldraw[fill=yellow!20!white, draw=yellow!40!black] (17,4) rectangle (18,7) (17,5) -- (18,5) (17,6) -- (18,6) ;
    \filldraw[fill=orange!20!white, draw=orange!40!black] (17,8) rectangle (18,11) (17,9) -- (18,9) (17,10) -- (18,10) ;
    \filldraw[fill=purple!20!white, draw=purple!40!black] (17,12) rectangle (18,15) (17,13) -- (18,13) (17,14) -- (18,14) ;
    \filldraw[fill=blue!20!white, draw=blue!40!black] (17,16) rectangle (18,19) (17,17) -- (18,17) (17,18) -- (18,18) ;
%    \filldraw[fill=green!20!white, draw=blue!40!black] (20,1) rectangle (21,17) (20,2) -- (21,2) (20,5) -- (21,5) (20,8) -- (21,8) (20,11) -- (21,11) (20,14) -- (21,14);
    \filldraw[fill=green!20!white, draw=green!40!black] (20,1) rectangle (21,2);
    \filldraw[fill=red!20!white, draw=red!40!black] (20,2) rectangle (21,5);
    \filldraw[fill=yellow!20!white, draw=yellow!40!black] (20,5) rectangle (21,8);
    \filldraw[fill=orange!20!white, draw=orange!40!black] (20,8) rectangle (21,11);
    \filldraw[fill=purple!20!white, draw=purple!40!black] (20,11) rectangle (21,14);
    \filldraw[fill=blue!20!white, draw=blue!40!black] (20,14) rectangle (21,17);
    \draw[->] (18,1.5) -- (20,3.5);
    \draw[->] (18,5.5) -- (20,6.5);
    \draw[->] (18,9.5) -- (20,9.5);
    \draw[->] (18,13.5) -- (20,12.5);
    \draw[->] (18,17.5) -- (20,15.5);
    \draw (1,-1.5) -- (19,-1.5);
    \draw[->] (19,-1.5) |- (20,1.5);
    \draw (23.5,11.5) node[text width=1cm] {relu}; %MLP Policy
    \draw (22.5,9.5) node [circle,draw,align=center,text width=0.2cm,radius=1] (z4) {} (22.2,9.2) -- (22.5,9.2) -- (22.7,9.8);
    \draw[->] (21,9.5) -- (z4);
    \draw[->] (z4) -- (24,9.5);
    \filldraw[fill=black!20!white, draw=black] (24,7.5) rectangle (25,11.5);
    \draw (27.5,11.5) node[text width=1cm] {relu};
    \draw (26.5,9.5) node [circle,draw,align=center,text width=0.2cm,radius=1] (z5) {} (26.2,9.2) -- (26.5,9.2) -- (26.7,9.8);
    \draw[->] (25,9.5) -- (z5);
    \draw[->] (z5) -- (28,9.5);
    \filldraw[fill=black!20!white, draw=black] (28,7.5) rectangle (29,11.5);
    \draw (31.5,11.5) node[text width=1cm] {linear};
    \draw (30.5,9.5) node [circle,draw,align=center,text width=0.2cm,radius=1] (z6) {}  (30.2,9.2) --  (30.7,9.7);
    \draw[->] (29,9.5) -- (z6);
    \draw[->] (z6) -- (32,9.5);
    \draw (35,9.5) node[text width=1cm] {policy}; %FINAL
    \filldraw[fill=black!20!white, draw=black] (32,9) rectangle (33,10);
    \draw (31.5,16) node[text width=1cm] {linear};
    \draw (30.5,14) node [circle,draw,align=center,text width=0.2cm,radius=1] (z3) {} (30.2,13.7) --  (30.7,14.2);
    \draw[->] (29,9.5) -- (z3);
    \draw[->] (z3) -- (32,14);
    \draw (35,14.2) node[text width=1cm] {value};
    \filldraw[fill=black!20!white, draw=black] (32,13.5) rectangle (33,14.5);
  \end{tikzpicture}
  \end{center}
  \caption[Architecture for our PPO neural network. ]{Architecture for our PPO neural network. An embedding generator calculates the embeddings for each word, and the mean of all word embeddings is used as the sentence embedding. The 6 coloured inputs are combined into a single layer of size 501 and then fed into 2 hidden layers of size 200 each. The outputs are used as the policy function and value function for PPO.}
  \label{fig:graphPPO}
\end{figure}

The neural network architecture for PPO is shown in Figure~\ref{fig:graphPPO}. As in~\cite{Molla:BioASQ7b},
the inputs of the neural network consist of:
\begin{enumerate}
    \item Candidate sentence
    \item Question
    \item Summary generated so far
    \item Sentences after the candidate sentence
    \item Entire document
    \item Length of summary generated so far
\end{enumerate}

For each of the first five inputs, we take the mean of the word2vec word embeddings (each of size 100) to generate the sentence embeddings. The sentence embeddings and the length of the summary are concatenated to form a single layer of size 501.
We then feed the combined layer into a simple Multi-Layer Perceptron with two hidden layers each of size 200.
The outputs of this neural network become the value function and stochastic policy function for our PPO approach, each of size 2 with linear activations, representing the predicted future rewards and action probabilities respectively for the two actions (classify~0 or~1).

\subsection{With BERT}\label{sec:ReBERT}
We also apply pre-trained BERT embeddings to our reinforcement learning approach for batches~4 and~5 of BioASQ8b.
Our network architecture is the same as in Figure~\ref{fig:graphPPO}, but we change the embedding generator to generate BERT embeddings instead of word2vec embeddings. %leaving the mean of embeddings as is
We use the PyTorch BERT embedding generator provided by Huggingface\textsuperscript{\ref{foot:huggingface}} and the TensorFlow Multi-Layer Perceptron policy provided by stable baselines\textsuperscript{\ref{foot:stable}}.% because the TensorFlow versions used by both libraries are not compatible.
%We receive a maximum ROUGE-SU4 F1 score of 0.2738 using BERT embeddings, where using word2vec embeddings only reaches 0.2691. (Reinforce: 0.2653 with 8b Data. Compare with 7b? Not much time to run multiple times after 8b data released)
We observe a minor improvement in ROUGE-SU4 F1 score as shown in Section~\ref{sec:submissions} (PPO BERT).

\section{Results and Discussion}\label{sec:results}

\subsection{Cross-Validation Results}

All experiments except those described in Sections~\ref{sec:RL} were evaluated using 10-fold cross-validation using the training data provided by the organisers of BioASQ8b. These results are shown in Table~\ref{tab:experiments}. As described in Section~\ref{sec:ReResults}, cross-validation evaluation was not practical for the experiments with reinforcement learning.

% Bar plots based on https://www.keithv.com/software/barchart/

  \tikzstyle barchart=[fill=black!20,draw=black]
  \tikzstyle errorbar=[very thin,draw=black!75]
  \tikzstyle sscale=[very thin,draw=black!75]

\newcommand{\mybar}[2]{
% #1 must be (value - 0.22) x 100
% #2 must be value x 100
  \begin{minipage}[c]{7.5cm}
   \begin{tikzpicture}
    \draw (0cm,0cm) (7.5,0.5);
  \draw[barchart] (0,0.152) rectangle (#1,0.438);
  \draw[errorbar] (#1-#2,0.295) -- (#1+#2,0.295);
  \draw[errorbar] (#1-#2,0.333) -- (#1-#2,0.258);
  \draw[errorbar] (#1+#2,0.333) -- (#1+#2,0.258);
   \end{tikzpicture}
  \end{minipage}
}

\newcommand{\mybarb}[2]{
% #1 must be (value - 0.22) x 100
% #2 must be value x 100
  \begin{minipage}[c]{7.5cm}
   \begin{tikzpicture}
    \draw (0cm,0cm) (7.5,0.5);
  \draw[barchart] (0,0.152) rectangle (#1,0.438);
  \draw[errorbar] (#1-#2,0.295) -- (7,0.295);
  \draw[errorbar] (#1-#2,0.333) -- (#1-#2,0.258);
   \end{tikzpicture}
  \end{minipage}
}

\newcommand{\mybarc}[2]{
% #1 must be (value - 0.22) x 100
% #2 must be value x 100
  \begin{minipage}[c]{7.5cm}
   \begin{tikzpicture}
    \draw (0cm,0cm) (7.5,0.5);
  \draw[barchart] (0,0.152) rectangle (#1,0.438);
  \draw[errorbar] (0,0.295) -- (#1+#2,0.295);
  \draw[errorbar] (#1+#2,0.333) -- (#1+#2,0.258);
   \end{tikzpicture}
  \end{minipage}
}

\newcommand{\myscale}{
\hspace{-3pt}%
 \begin{minipage}[c]{7cm}
   \begin{tikzpicture}
    \draw (0cm,0cm) (6,0.3);
    \draw[sscale] (0,0.3) -- (6,0.3);
    \draw[sscale] (0,0.3) -- (0,0.4);
    \draw[sscale] (0,0) node[text width=0pt, text height=0pt, font=\footnotesize] {0.23};
    \draw[sscale] (1,0.3) -- (1,0.4);
    \draw[sscale] (1,0) node[text width=0pt, text height=0pt, font=\footnotesize] {0.24};
    \draw[sscale] (2,0.3) -- (2,0.4);
    \draw[sscale] (2,0) node[text width=0pt, text height=0pt, font=\footnotesize] {0.25};
    \draw[sscale] (3,0.3) -- (3,0.4);
    \draw[sscale] (3,0) node[text width=0pt, text height=0pt, font=\footnotesize] {0.26};
    \draw[sscale] (4,0.3) -- (4,0.4);
    \draw[sscale] (4,0) node[text width=0pt, text height=0pt, font=\footnotesize] {0.27};
    \draw[sscale] (5,0.3) -- (5,0.4);
    \draw[sscale] (5,0) node[text width=0pt, text height=0pt, font=\footnotesize] {0.28};
    \draw[sscale] (6,0.3) -- (6,0.4);
%    \draw[sscale] (6,0) node[text width=0pt, text height=0pt, font=\footnotesize] {0.29};
%    \draw[sscale] (7,0.3) -- (7,0.4);
%    \draw[sscale] (7,0) node[text width=0pt, text height=0pt, font=\footnotesize] {0.30};
   \end{tikzpicture}
  \end{minipage} 
  }

\begin{table}
    \centering
    \caption{Results of experiments using
      ROUGE SU4 F-score under 10-fold cross-validation. The table shows the mean
      and standard deviation across the folds. ``firstn'' is a baseline that selects the first n sentences. NNR and NNC are described in Section~\ref{sec:framework}. BERT and BioBERT are described in Section~\ref{sec:BERT}. Siamese LSTM is described in Section~\ref{sec:SiamLSTM}. The SBERT and SBioBERT runs are described in Section~\ref{sec:SiamBERT}.}
    \label{tab:experiments}
    \begin{tabular}{lr@{ $\pm$ }ll}
    %\toprule
         \textbf{Method} & \multicolumn{2}{c}{\textbf{ROUGE-SU4 F1}} \\
                & \textbf{Mean} & \textbf{1 stdev}\\
         \cmidrule{1-3}
      firstn & 0.261 &  0.011 & \mybar{3.1}{1.1}\\
      \cmidrule{1-3}
      NNR & 0.264 & 0.008 & \mybar{3.4}{0.8}\\
      NNC & 0.271 & 0.013 & \mybar{4.1}{1.3}\\
      \cmidrule{1-3}
      BERT untrained & 0.270 & 0.014 & \mybar{4.0}{1.4}\\
      BERT trained & 0.261 & 0.012 & \mybar{3.1}{1.2}\\
      BERT LSTM & 0.274 & 0.010 & \mybar{4.4}{1.0}\\
      \cmidrule{1-3}
      BioBERT untrained & 0.262 & 0.010 & \mybar{3.2}{1.0}\\
      BioBERT LSTM & 0.264 & 0.012& \mybar{3.4}{1.2} \\
      \cmidrule{1-3}
      Siamese LSTM & 0.263 & 0.010 & \mybar{3.3}{1.0}\\
      \cmidrule{1-3}
%      sBERT R & 0.289 & 0.022 & \mybarb{5.9}{2.2}\\
%      sBERT C & 0.243 & 0.022 & \mybarc{1.3}{2.2}\\
%      sBERT M R & 0.293 & 0.025 & \mybarb{6.3}{2.5}\\
%      sBERT M C & 0.254 & 0.022 & \mybar{2.4}{2.2}\\
%      \cmidrule{1-3}
%      sBioBERT R & 0.286 & 0.023 & \mybarb{5.6}{2.3}\\
%      sBioBERT C & 0.253 & 0.022 & \mybar{2.3}{2.2}\\
%      sBioBERT M R & \textbf{0.298} & 0.024 & \mybarb{6.8}{2.4}\\
%      sBioBERT M C & 0.257 & 0.023 & \mybar{2.7}{2.3}\\
      \cmidrule{1-3}
 \multicolumn{3}{l}{} & \myscale\\
 %\bottomrule
    \end{tabular}
\end{table}

%\begin{table}[]
%\begin{tabular}{lp{8cm}r}
%\textbf{Run}              & \textbf{Description}                                                       & ROUGE-SU4 F1                              \\ \hline
%sbert reg                 & Siamese BERT with Regression (1 epoch)                                     & 0.28923 $\pm$ 0.02254                     \\
%sbert cls                 & Siamese BERT with Classfication (1 epoch)                                  & 0.24309 $\pm$ 0.02241                     \\
%multitask sbert reg       & Siamese BERT multitask, with regression prediction layer (1 epoch)         & \textbf{0.29324 $\pm$ 0.02494}                     \\
%multitask sbert cls       & Siamese BERT multitask, with classification prediction layer  (1 epoch)    & 0.25356 $\pm$ 0.02206                     \\
%sbert reg (bio)           & Siamese BioBERT with Regression (1 epoch)                                  & \multicolumn{1}{l}{0.28680 $\pm$ 0.02334} \\
%sbert cls (bio)           & Siamese BioBERT with Classfication (1 epoch)                               & \multicolumn{1}{l}{0.25310 $\pm$ 0.02197} \\
%multitask sbert reg (bio) & Siamese BioBERT multitask, with regression prediction layer (1 epoch)      & \multicolumn{1}{l}{\textbf{0.29798 $\pm$ 0.02399}} \\
%multitask sbert cls (bio) & Siamese BioBERT multitask, with classification prediction %layer  (1 epoch) & \multicolumn{1}{l}{0.25708 $\pm$ 0.02261}
%\end{tabular}
%\end{table}

The baseline runs ``firstn'', ``NNR'' and ``NNC'' confirm~\cite{Molla:BioASQ7b}'s observation that the classification setup produces better results than the regression setup. The untrained BERT system did not better the classification system, and the trained BERT system produced worse results.

We observed no changes in the evaluation results of the BERT trained system as we changed the number of epochs from 1 to 20 epochs. A detailed look at the changes in the loss during training revealed a very small improvement of the loss as we increase the number of epochs, but not enough to reflect a difference in the final ROUGE-SU4 F1 metric. This suggests that a more elaborate fine-tuning regime with gradual unfreezing as described by~\cite{howard:ruder:2018} might lead to improved results.

Given the poor results of the trained BERT system, we kept BERT frozen when we tested the variants using LSTM and using BioBERT.

BERT and BioBERT followed by an LSTM-based sentence reductor did improve results over the version with a mean of embeddings. Unfortunately, the experiments using the LSTM-based reductor were made after the deadlines for submission of results to BioASQ8b.

It was surprising to observe that the BioBERT variants performed worse than the BERT variants. This is not in line with the improvement in the performance of BioBERT for the ``exact answers'' section of BioASQ7b reported in literature~\cite{Yoon:2019}.

The Siamese LSTM variant did not improve over the classification system. The reason for this might be, as mentioned in Section~\ref{sec:submissions}, that questions and candidate sentences are different in nature. We also conducted cross-validation experiments with the SBERT variants but they are not included in Table~\ref{tab:experiments} as the results are very different from those of the runs submitted to BioASQ (Section~\ref{sec:results}) and we suspect that there might have been an error when running these evaluations.
%the BERT and BioBERT variants showed a clear improvement in our experiments. It was interesting to observe that the regression settings of the Siamese Bert and BioBERT had much better results than the classification settings. On the other hand, the multi-task versions that incorporated regression and classification had better results than the regression systems. These results, however, do not concur with the results of the runs submitted to BioASQ8b~\ref{sec:results} and we suspect that there might have been an error when running our experiments.

Table~\ref{tab:hyperparameters} shows and explains the hyperparameters of all systems of Table~\ref{tab:hyperparameters}.

\begin{table}
    \centering
    \caption{Hyperparameters for the experiments of Table~\ref{tab:experiments}. The choice of dropout and epochs is the result of grid search. A dropout of 0 means that no dropout was applied. The choice of batch size is determined by the GPU capabilities. In all cases, the dimensions of the word and sentence embeddings is 100. The size of the LSTM states and intermediate layers is also 100. The number of candidate sentences is limited to 50. The BERT version is bert-base-uncased provided by \protect\url{https://huggingface.co/}. The BioBERT version is v1.1-pubmed available at \protect\url{https://github.com/dmis-lab/biobert}}
    \label{tab:hyperparameters}
    \begin{tabular}{lrrrr}
    \toprule
    \textbf{Method} & \textbf{Batch size} & \textbf{Dropout} & \textbf{Epochs} & \textbf{Sentence length}\\
    \midrule
NNR& 	1024 &	0.3 &	10 & 300	\\
NNC &	1024 &	0.3 &	10 & 300	\\
\midrule
BERT untrained &	32 &	0 &	50 & 250	\\
BERT trained &	8 &	0 &	1 & 250	\\
BERT LSTM &	1024 &	0.6 &	10 & 250	\\
\midrule
BioBERT untrained &	1024 &	0 &	10 & 250	\\
BioBERT LSTM &	1024 &	0.6 &	10 & 250	\\
\midrule
Siamese LSTM &	1024 &	0.2 &	10 & 300	\\
    \bottomrule
    \end{tabular}
\end{table}

\subsection{Reinforcement Learning Results}\label{sec:ReResults}

The evaluation setup for reinforcement learning did not use cross-validation due to the long time it took to train the system (several days). We therefore used a partition of the training data for the train and test sets.

Table~\ref{tab:ReinforceResults} shows the evaluation results of the PPO \cite{schulman:2017} reinforcement learning approaches discussed in Section~\ref{sec:RL}, compared with the REINFORCE \cite{williams:1992} approach, using the BioASQ7b and BioASQ8b training datasets. %Appendix~\ref{app:ppo} shows the learning curves for the PPO runs submitted to BioASQ8b.

\begin{table}
  \centering
  \caption{Reinforcement Learning Preliminary Results. The results using the BioASQ7b dataset show the mean and standard deviation of 3 runs. The results using the BioASQ8b dataset show the result of a single run. The PPO BERT system was only run with the BioASQ8b dataset and was only submitted to batches~4 and~5 of BioASQ8b. All results use the ROUGE-SU4 F1 metric.}
  \label{tab:ReinforceResults}
  \begin{tabular}{p{3.5cm}r@{ $\pm$ }lr}
  \toprule
    %System & \multicolumn{3}{c}{Dataset (ROUGE-SU4 F1)}\\
    System & \multicolumn{2}{l}{BioASQ7b} & \multicolumn{1}{l}{BioASQ8b} \\
     & Mean & stdev & \\
    \midrule
    REINFORCE word2vec & 0.253 & 0.001 & 0.265 \\
    PPO word2vec & 0.251 & 0.001 & 0.269 \\
    PPO BERT & \multicolumn{2}{c}{-} & 0.274 \\
    \bottomrule
  \end{tabular}
\end{table}

The \textbf{BioASQ7b} data set consists of 2,747 questions which we divide into a training set (2,289 questions) and a testing set (458 questions) partitioned 5:1 using a random seed to shuffle the data.
The PPO word2vec and REINFORCE word2vec systems were run 3 times each for 500,000 timesteps on the BioASQ7b data set, and the maximum ROUGE-SU4 F1 score reached in each learning curve was averaged across the 3 runs.
%As we report in \cite{jones:2020}, the difference between the results for PPO and REINFORCE is not significant, but the learning curve for PPO is more consistent and does not change suddenly.

The \textbf{BioASQ8b} data set consists of 3,243 questions which we divide into a training set (2,702 questions) and a testing set (541 questions) partitioned 5:1 using a random seed to shuffle the data.
The PPO word2vec, PPO BERT, and REINFORCE word2vec systems were each run once only for 500,000 timesteps on the BioASQ8b data set, and the maximum ROUGE-SU4 F1 score for each run is shown in Table~\ref{tab:ReinforceResults}.
The models which produced the maximum ROUGE-SU4 F1 score for each system were saved and re-used to generate our PPO submissions to the BioASQ competition in 2020, except for the REINFORCE system which was not included in this year's submission.

Whereas our evaluation results did not show any clear difference between REINFORCE and PPO on word2vec embeddings, the version using PPO and BERT features showed an improvement.

The reinforcement learning experiments of Table~\ref{tab:ReinforceResults}, however, do not outperform the results of the experiments of Table~\ref{tab:experiments}.

%for PPO using ROUGE-SU4 evaluation metrics do not outperform REINFORCE for this task, and we are keen to find out the human evaluations for our submissions to BioASQ8b competition in 2020.

%We observed minor improvements in our results when using BERT embeddings instead of word2vec embeddings.
%However, all improvements which we observed constitute small changes in ROUGE-SU4 F1 score which are not very significant.

\section{Submissions to BioASQ8b}\label{sec:submissions}

Table~\ref{tab:bioasq8b} shows the results of \institutiona's submissions to BioASQ8b. We only report ROUGE-SU4 F1 for comparison with our experiments. Also, as observed by~\cite{Molla:BioASQ7b}, F1 scores have a higher correlation with human judgements than recall scores, and similar correlation with human judgements as precision scores.

\begin{table}
  \centering
  \caption{MQ runs submitted to BioASQ8b}
  \label{tab:bioasq8b}
  \begin{tabular}{lllr}
  \toprule
    \textbf{Batch} & \textbf{Run Name} & \textbf{Description} & \textbf{ROUGE-SU4 F1}\\
    \midrule
    1 & MQ1 & First $n$ & 0.3302\\
          & MQ2 & NNR batchsize=4096 & 0.3508\\
          & MQ3 & NNC batchsize=4096 & \textbf{0.3556}\\
          & MQ4 & BERT untrained & 0.3359\\
          & MQ5 & PPO word2vec & 0.3386\\
    \midrule
    2 & MQ1 & First $n_b$ & 0.2897\\
          & MQ2 & NNR batchsize=4096 & 0.3360\\
          & MQ3 & NNC batchsize=4096 & \textbf{0.3376}\\
          & MQ4 & BERT untrained & 0.3030\\
          & MQ5 & PPO word2vec & 0.2662\\
    \midrule
    3 & MQ1 & First $n$ & 0.3506\\
          & MQ2 & NNR batchsize=4096 & \textbf{0.3729}\\
          & MQ3 & NNC batchsize=4096 & 0.3651\\
          & MQ4 & BERT untrained & 0.3506\\
          & MQ5 & PPO word2vec & 0.3018\\
    \midrule
    4 & MQ1 & PPO BERT & 0.2975\\
          & MQ2 & NNR batchsize=4096 & 0.2973\\
          & MQ3 & NNC batchsize=4096 & \textbf{0.2987}\\
          & MQ4 & BERT untrained & 0.2954\\
          & MQ5 & PPO word2vec & 0.2585\\
    \midrule
    5 & MQ1 & PPO BERT & 0.3135\\
          & MQ2 & NNR batchsize=4096 & 0.3237\\
          & MQ3 & NNC batchsize=4096 & 0.3276\\
          & MQ4 & BERT untrained & \textbf{0.3316}\\
          & MQ5 & PPO word2vec & 0.3096\\
    \bottomrule
  \end{tabular}
\end{table}

All runs in Table~\ref{tab:bioasq8b} have been described except for First $n_b$ in batch~2. This is the same as First $n$ in all other batches but the data has been pre-processed differently. 

The classifier system (NNC) produces the best results in most runs, and this is consistent with our cross-validation results (Table~\ref{tab:experiments}). However, in contrast with Table~\ref{tab:experiments}, in most runs the second best is the regression system (NNR) instead of the untrained BERT system. All systems were re-trained for the BioASQ runs, keeping the same hyperparameters, so that we could use the entire training data, and it is possible that bad luck played a part here, and the BERT system was not trained to its best.

The reinforcement learning experiments show the lowest evaluation results. We should note, however, that the RL runs submitted to BioASQ7b also had lower results than the other runs, but the human evaluation results ranked them higher than our other runs. Also, the reinforcement learning submissions were not re-trained using the entire training data, which may be worth exploring because PPO had a more consistent learning curve for us than REINFORCE. We are waiting for the human evaluation results of BioASQ8b with anticipation. 

Table~\ref{tab:sbert_runs} shows the results of \institutionb's submissions to BioASQ8b.
\begin{table}[]
\centering
\caption{SBERT runs submitted to BioASQ8b}\label{tab:sbert_runs}
\begin{tabular}{lllr}
\toprule
\textbf{Batch} & \textbf{Run Name}        & \textbf{Description}                                                    & \textbf{ROUGE-SU4 F1}    \\
\midrule
2              & sbert reg           & SBERT R 1 epoch                                  & 0.2856          \\
               & sbert cls           & SBERT C 1 epoch                               & \textbf{0.2958} \\
               & multitask sbert reg & SBERT M R 1 epoch      & 0.2857          \\
               & multitask sbert cls & SBERT M C 1 epoch & 0.2910          \\
\midrule
3              & sbert reg           & SBERT R 20 epochs                                                                & 0.3241          \\
               & sbert cls           & SBERT C 20 epochs                                                                & 0.3349          \\
               & multitask sbert reg & SBERT M R 20 epochs                                                                & 0.3342          \\
               & multitask sbert cls & SBERT M C 20 epochs                                                                & \textbf{0.3506} \\
               & sbert 1 epoch cls   & SBERT C 1 epoch                                                                 & 0.3395          \\
\midrule
4              & sbert reg           & SBioBERT R 20 epochs                                                        & \textbf{0.2601} \\
               & sbert cls           & SBioBERT C 20 epochs                                                        & \textbf{0.2601} \\
               & multitask sbert reg & SBioBERT M R 20 epochs                                                        & 0.2572          \\
               & multitask sbert cls & SBioBERT M C 20 epochs                                                        & 0.2527          \\
               & sbert cls 1 epoch   & SBioBERT C 1 epoch                                                         & 0.2580          \\
\midrule
5              & sbert reg           & SBioBERT R 20 epochs                                                      & 0.3201          \\
               & sbert cls           & SBioBERT C 20 epochs                                                        & 0.3235          \\
               & multitask sbert reg & SBioBERT M R 20 epochs                                                       & 0.3152          \\
               & multitask sbert cls & SBioBERT M C 20 epochs                                                       & \textbf{0.3245}          \\
               & sbert cls 1 epoch   & SBioBERT C 1 epoch                                                        & 0.3231          \\
\bottomrule
\end{tabular}
\end{table}

We observe that the SBERT runs perform worse or on par with the BERT untrained model of Table~\ref{tab:bioasq8b}. This might be because questions and answers may not appear closely in an embedding space, and therefore an identical processing of each of them might not be advantageous. For example, a question includes question words whereas the candidate sentences are not normally questions. We also observe that the BioBERT models seem to produce lower results when compared with the BERT untrained model of Table~\ref{tab:bioasq8b}. In general, the multitask learning approach performed better than just classification or regression.
%    1.	Siamese BERT performed worse or on par with a BERT model (comparing with your runs).
%    This might be because questions and answers do not appear closely in an embedding space (a question probably doesn’t have the same semantic content as the answer, otherwise why ask the question)
%    2.	After I switched to BioBERT model (round 4 and 5), my runs dropped in performance
%    3.	Multitask learning performed better than just classification or regression.

\section{Conclusions}\label{sec:conclusions}

This paper presents our approaches to query focused multi-document extractive summarisation for BioASQ8b. Our experiments include the use of BERT and BioBERT, Siamese architectures and SBERT, and Reinforcement Learning with PPO.

We observed that an approach that uses BERT to obtain the word embeddings, followed by LSTM to map these word embeddings to sentence embeddings, had the most promising results. The variant with BioBERT did not present an improvement, and this conflicts with the overall improvement of the use of BioBERT for question answering on biomedical texts. The approaches with Siamese architectures did not present an improvement over the base versions, presumably because questions and candidate sentences have different sentence styles.

As further research we plan to explore fine-tuning techniques for the BERT and BioBERT variants.

\section*{Acknowledgements}

Research by Vincent Nguyen is supported by the Australian Research Training Program and the CSIRO Postgraduate Scholarship.

%\appendix

%\section{Appendix: PPO Learning Curves BioASQ8b}\label{app:ppo} %can remove if we don't think we need...

%%\begin{figure}
%  \begin{center}
%  \includegraphics[width=0.7\linewidth]{RLLearning.jpg}
%  \end{center}
%%  \caption[]{ Appendix: PPO Learning Curves }
%%  \label{fig:xxx}
%%\end{figure}
%The learning curves when training the PPO word2vec and PPO BERT models on the BioASQ 8b training data.

\bibliographystyle{splncs04}
\bibliography{mybibliography}

\end{document}